\documentclass{article}
\usepackage{spconf,amsmath,epsfig}
\usepackage[pagebackref,breaklinks,colorlinks]{hyperref}

\title{Smartflow: Enabling Scalable Spatiotemporal Geospatial Research}
\name{David McVicar, Brian Avant, Adrian Gould, Diego Torrejon, Charles Della Porta, Ryan Mukherjee}
\address{BlackSky}

\begin{document}
%
\maketitle
%


\begin{abstract}
BlackSky introduces Smartflow, a cloud-based framework enabling scalable spatiotemporal geospatial research built on open-source tools and technologies. Using STAC-compliant catalogs as a common input, heterogeneous geospatial data can be processed into standardized datacubes for analysis and model training. Model experimentation is managed using a combination of tools, including ClearML, Tensorboard, and Apache Superset. Underpinning Smartflow is Kubernetes, which orchestrates the provisioning and execution of workflows to support both horizontal and vertical scalability. This combination of features makes Smartflow well-suited for geospatial model development and analysis over large geographic areas, time scales, and expansive image archives.

We also present a novel neural architecture, built using Smartflow, to monitor large geographic areas for heavy construction. Qualitative results based on data from the IARPA Space-based Machine Automated Recognition Technique (SMART) program\footnote{\small \url{https://www.iarpa.gov/research-programs/smart}} are presented that show the model is capable of detecting heavy construction throughout all major phases of development.

\end{abstract}

\section{Introduction}

BlackSky recently added two new satellites to its constellation, providing affordable access to high-revisit satellite imagery~\cite{bksycommissioning}. BlackSky's first-of-a-kind commercial constellation provides reliable and dynamic hourly monitoring, up to 15 times a day from dawn until dusk. More generally, there are significantly more earth observation satellites in operation today compared to only a few years ago. These satellites are enabling a wide variety of applications, including intelligence, socioeconomic monitoring, addressing climate change, and more~\cite{fmow2018, hadzic2020estimating, jean2016combining, climate2021}. 

As the quantity of data and number of potential applications increase, it is becoming evident that automated systems are required to efficiently derive insights for human analysts and decision makers. Efficiency is especially critical given that historical data archives are already at multi-petabyte scale~\cite{gorelick2017google}. Furthermore, the scale of available data makes it near impossible for human analysts to fully examine all historical and multi-modal data for an area of interest, implying that many applications may only be possible via automated techniques~\cite{erwin2012too}. The system we discuss in this paper seeks to address these issues in order to enable new geospatial applications.

\begin{keywords}
Smartflow, geospatial, satellite imagery, deep learning, AI, spatiotemporal, IARPA SMART, construction, change detection
\end{keywords}

\section{Smartflow}

\begin{figure*}[htb]
\begin{minipage}[b]{1.0\linewidth}
  \centering
\centerline{\epsfig{figure=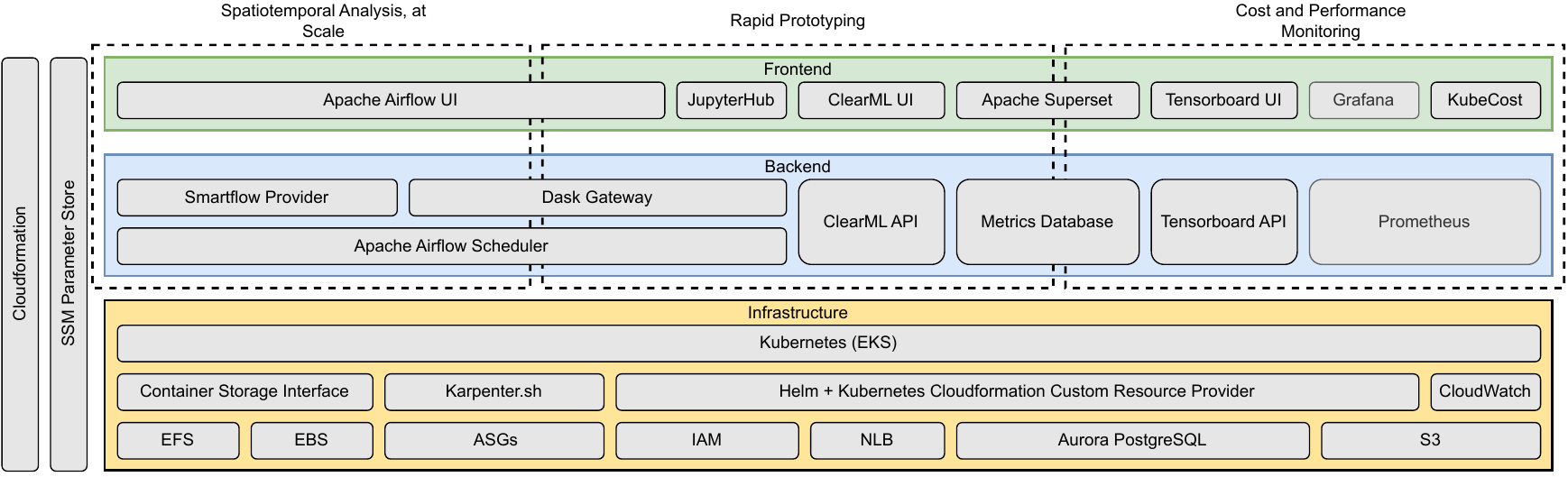,width=17cm}}\medskip
\end{minipage}
\begin{minipage}[b]{1.0\linewidth}
  \centering
\centerline{\epsfig{figure=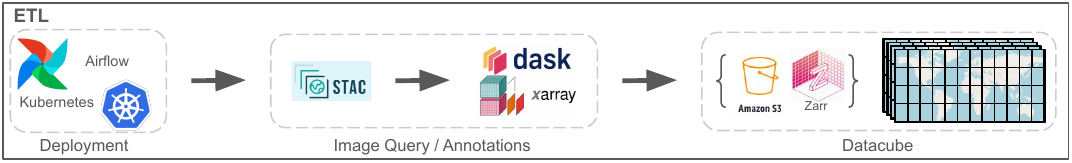,width=17.0cm}}\medskip
\end{minipage}
\begin{minipage}[b]{1.0\linewidth}
  \centering
\centerline{\epsfig{figure=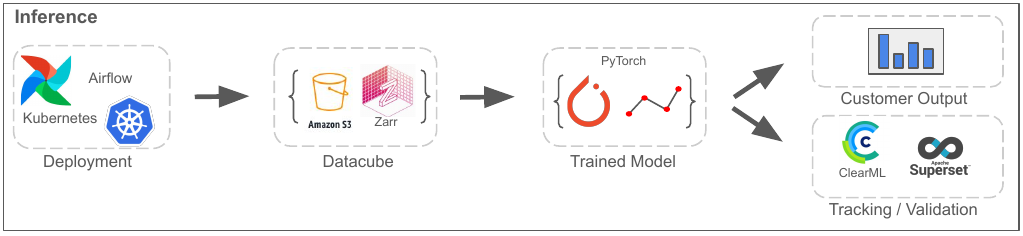,width=17.0cm}}
\end{minipage}
\caption{Diagram of components in the Smartflow framework.}
\label{fig:smartflow}
\end{figure*}

Granular AI's GeoEngine is a recent attempt at solving the aforementioned issues and is also the most similar published work to ours~\cite{verma2022geoengine}. GeoEngine assembles a mix of proprietary and open-source tools into an architecture for supporting model training and experimentation, as well as model inference and deployment. Our system, named ``Smartflow'', offers similar functionality to GeoEngine but is built exclusively on open-source tools, which bring a variety of benefits including flexibility, cost savings, and quality~\cite{khan2012free}. Additionally, Smartflow leverages the latest developments in the Pangeo geoscience community, such as STAC-based data management, Xarray, Dask, and Zarr~\cite{hamman2018pangeo}.

In Smartflow, rather than limiting users to manually ingested and catalogued data, tools like PySTAC\footnote{\small \url{https://github.com/stac-utils/pystac}} are used to query and source data from STAC-compliant catalogs. Satellite imagery stored using the STAC specification\footnote{\small \url{https://stacspec.org}} enables rapid access to metadata which can then be used to lazily query image pixels. This lazy access process reduces unnecessary data transfers and geospatial transformations.

Combining Dask's delayed computations and efficient storage with Zarr, Smartflow offers an extract-transform-load (ETL) pipeline to efficiently retrieve, transform, and store collections of spatiotemporal assets across heterogeneous sensors. Items retrieved from STAC queries can be chunked by dimension (i.e., spatial, temporal, modality) and operations can be applied over these dimensions using Xarray's API. Further, when combined with Dask, array operations can be delayed and mapped in a task graph which can be deployed across a distributed cluster of workers to maximize efficiency.

Beyond ETL, Smartflow serves as a platform for spatiotemporal analysis at scale, rapid prototyping, and cost and performance monitoring as shown in Figure~\ref{fig:smartflow}. At the core of the Smartflow framework, Kubernetes orchestrates the provisioning and execution of workflows in the cloud. Kubernetes allows workflows to scale horizontally and vertically, allowing for flexible infrastructure that is specifically provisioned for each workflow. These workflows are initialized and monitored through Apache Airflow, with custom operators to support processing arbitrary geographic regions. Dask Gateway is utilized to provision parallel computing workers to execute MapReduce-style tasks. This allows workflows to process large quantities of data on commodity, low-cost hardware. Smartflow leverages Tensorboard and the open-source tier of ClearML for experiment management and tracking. It also provides support for Apache Superset, allowing users to visualize metrics stored in virtually any RDBMS. A variety of metrics across the stack are also tracked via Prometheus and are easily visualized via Grafana. With these tools, Smartflow can efficiently deploy geospatial workloads to any scale, even global.

Smartflow is deployed via CloudFormation and is triggered via an easy-to-use python script. This allows multiple Smartflow environments to be created and used in parallel, each with their own set of security and scalability parameters, enabling isolated and appropriately-provisioned development and test environments.

With this framework, BlackSky won a competitive award to provide the core infrastructure to the entire IARPA SMART program \cite{bksyframework}.

\section{Construction Model}

\begin{figure*}[htb]
\begin{minipage}[b]{1.0\linewidth}
  \centering
  \centerline{\epsfig{figure=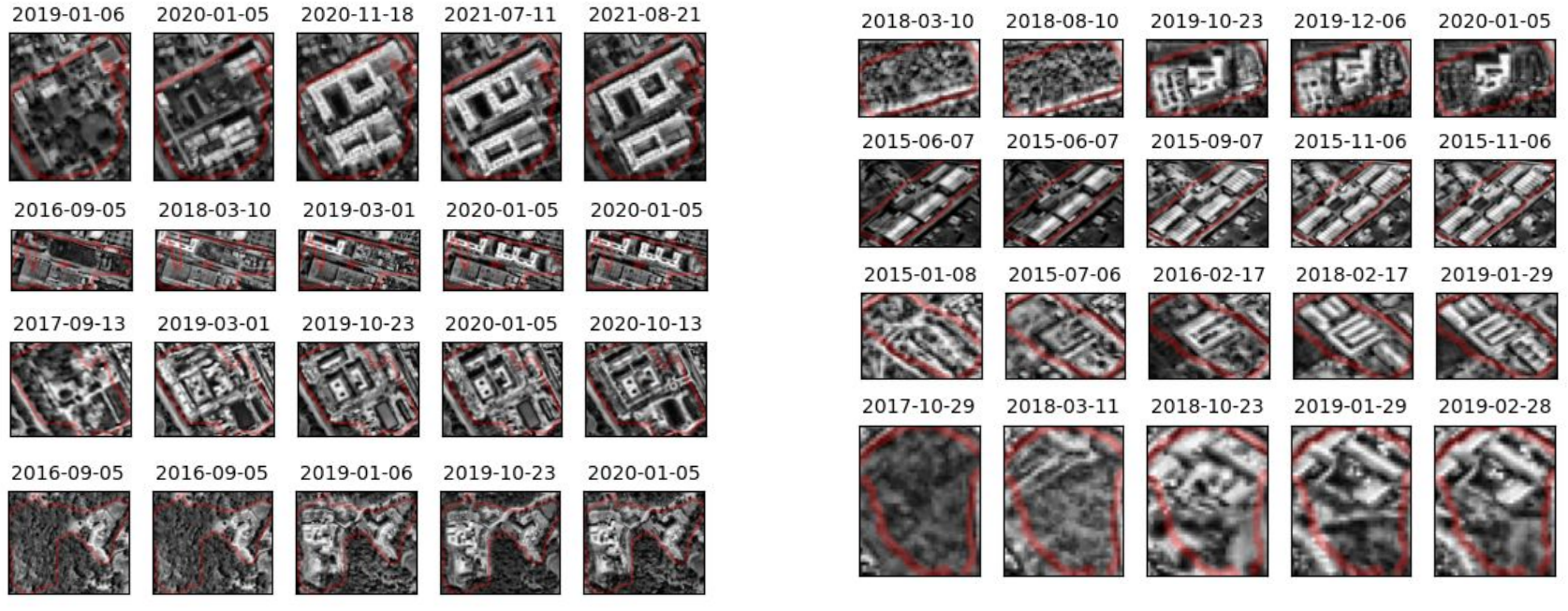,width=17cm}}
\end{minipage}
\caption{Qualitative heavy construction results from Jacksonville, FL and Christchurch, NZ. Eight sequences of five images each are shown. Our model operates on sequences of ten, but here we select a subset for compactness. A red outline in each sequence captures the construction site shape estimated by our model.}
\label{fig:results}
\end{figure*}

Using Smartflow, we developed a novel deep-learning model to identify heavy construction occurring over broad geographic areas, time scales, and heterogeneous sensor data, including Sentinel-2, Landsat, WorldView, and Planet. The model relies on spatially-aligned cubes of data generated using the Smartflow ETL process, aggregating imagery across time and sensor for each user-defined geographic region while using quality mask information to filter out imagery with artifacts and high presence of clouds. Once data cubes are generated, the model processes randomized temporal subsets from these cubes including information from the beginning, middle, and end of a time series in order to localize construction.

Annotations for training and evaluating models on this task were created by the IARPA SMART program and are detailed by Goldberg et al.~\cite{goldberg2023automated}. Unlike previous datasets, the annotations and evaluation methodology used here do not operate on a pre-selected set of imagery, but rather require systems to operate as they would in a production setting. Given only a geographic region and time frame, systems must search for and curate their own imagery for both training and inference, handling any issues that may arise due to sensor noise, clouds, environmental conditions, etc. Additionally, the target construction class is defined more narrowly than in other datasets, introducing an additional challenge. For example, large industrial and commercial construction is considered positive whereas the construction of single-family homes, sports fields, or road infrastructure are negative. As one can imagine, separating these different types of construction necessitates processing and understanding multiple temporal views.

In order to simplify the learning task, speed up training, and reduce computational complexity, our model architecture follows a spatiotemporal factorization by decoupling the mixing of features between space and time domains as explored by \cite{tarasiou2023vits}. In particular, images in the sequence are first processed independently by a U-Net \cite{ronneberger2015u} with an EfficientNet-B0 \cite{tan2019efficientnet} backbone. Representations from this step are intended to capture per-pixel spatial features relevant to identifying the location of construction sites. These spatial representations are then mixed using a subsequent U-Net to extract pixel-wise temporal information to localize the temporal extent of construction sites. Finally, spatiotemporal outputs are fed into a convolutional segmentation head to generate pixel-wise construction probabilities per time frame.

Qualitative results from this model, trained on data made available as part of the IARPA SMART program, are shown in Figure~\ref{fig:results}. In the figure, a sequence of 5 images is shown for each construction site with the detected construction boundaries outlined in red. It is apparent from these results that the model is not only capable of detecting active construction, but also of detecting other construction phases such as site preparation or land clearing. By looking at multiple subsets of time frames over long periods, the model is able to discern accurate and detailed site boundaries at all points in time. Extensions to this model have also been trained to classify specific phases of construction based on metadata derived from each of the detected sites. 

Quantitative metrics may be released at a later date conditioned on the release of program annotations. However, we believe our approach has proven effective at efficiently sifting through large quantities of data to locate heavy construction, even when it may be difficult for humans to identity.

\section{Conclusion}
We introduced Smartflow, a novel framework that enables scalable spatiotemporal geospatial research built upon open-source tools and cloud computing. We have shown how Smartflow can efficiently process large quantities of multi-temporal multi-sensor data using STAC-compliant catalogs and Kubernetes. We have also shown how Smartflow can facilitate global algorithm implementation, which can improve the generalizability and utility of deep learning models for geospatial applications. As a case study, we presented a new architecture for monitoring heavy construction across heterogeneous satellite data using Smartflow. Our framework offers a new ability to conduct geospatial research at unprecedented scales and resolutions, opening up new possibilities for a variety of applications. We plan to continue to refine Smartflow and extend our deep learning models to new tasks.

\subsection*{Acknowledgements}
This research is based upon work supported in part by the Office of the Director of National Intelligence (ODNI), Intelligence Advanced Research Projects Activity (IARPA), via contract 2021-20111000003. The views and conclusions contained herein are those of the authors and should not be interpreted as necessarily representing the official policies, either expressed or implied, of ODNI, IARPA, or the U.S. Government. The U.S. Government is authorized to reproduce and distribute reprints for governmental purposes notwithstanding any copyright annotation therein.

\bibliographystyle{IEEEbib}
\bibliography{refs}

\end{document}